Original

# Revolutionizing Blood Banks: AI-Driven Fingerprint-Blood Group Correlation for Enhanced Safety

Revolucionando los Bancos de Sangre: Correlación entre Huellas Dactilares y Grupos Sanguíneos Impulsada por IA para una Mayor Seguridad

Malik A. Altayar[1], maltayar@ut.edu.sa, https://orcid.org/0000-0002-3394-0686
Muhyeeddin Alqaraleh[2], malqaraleh@zu.edu.jo, https://orcid.org/0009-0001-9103-2002
Mowafaq Salem Alzboon[3], malzboon@jadara.edu.jo, https://orcid.org/0000-0002-3522-6689
Wesam T. Almagharbeh[4], Walmagharbeh@ut.edu.sa, https://orcid.org/0000-0002-8435-1208

[1] University of Tabuk, Faculty of Applied Medical Sciences, Tabuk, Saudi Arabia.
[2] Zarqa University, Faculty of Information Technology, Zarqa, Jordan.
[3] Jadara University, Faculty of Information Technology, Irbid, Jordan.
[4] University of Tabuk, Faculty of Nursing, Tabuk, Saudi Arabia.

**ABSTRACT**

Identification of a person is central in forensic science, security, and healthcare. Methods such as iris scanning and genomic profiling are more accurate but expensive, time-consuming, and more difficult to implement. This study focuses on the relationship between the fingerprint patterns and the ABO blood group as a biometric identification tool. A total of 200 subjects were included in the study, and fingerprint types (loops, whorls, and arches) and blood groups were compared. Associations were evaluated with statistical tests, including chi-square and Pearson correlation.

The study found that the loops were the most common fingerprint pattern and the O+ blood group was the most prevalent. Discussion: Even though there was some associative pattern, there was no statistically significant difference in the fingerprint patterns of different blood groups. Overall, the results indicate that blood group data do not significantly improve personal identification when used in conjunction with fingerprinting.

Although the study shows weak correlation, it may emphasize the efforts of multi-modal based biometric systems in enhancing the current biometric systems. Future studies may focus on larger and more diverse samples, and possibly machine learning and additional biometrics to improve identification methods. This study addresses an element of the ever-changing nature of the fields of forensic science and biometric identification, highlighting the importance of resilient analytical methods for personal identification.

**RESUMEN**

La identificación de una persona es fundamental en la ciencia forense, la seguridad y la atención médica. Métodos como el escaneo del iris y el perfilado genómico son más precisos, pero también costosos, requieren mucho tiempo y son más difíciles de implementar. Este estudio se centra en la relación entre los patrones de huellas dactilares y el grupo sanguíneo ABO como una herramienta de identificación biométrica. Se incluyeron un total de 200 sujetos en el estudio, y se compararon los tipos de huellas dactilares (lazos, espirales y arcos) con los grupos sanguíneos. Se evaluaron asociaciones mediante pruebas estadísticas, incluyendo chi-cuadrado y coeficiente de correlación de Pearson.

El estudio encontró que los lazos fueron el patrón de huella dactilar más común y que el grupo sanguíneo O+ fue el más prevalente. Discusión: Aunque se observaron algunos patrones asociativos, no hubo una diferencia estadísticamente significativa entre los patrones de huellas dactilares y los diferentes grupos sanguíneos. En general, los resultados indican que los datos de los grupos sanguíneos no mejoran significativamente la identificación personal cuando se combinan con la huella dactilar.

Aunque el estudio muestra una correlación débil, resalta la importancia de los sistemas biométricos multimodales en la mejora de los sistemas biométricos actuales. Estudios futuros podrían enfocarse en muestras más grandes y diversas, así como en el uso de aprendizaje automático y biometría adicional para mejorar los métodos de identificación. Este estudio aborda un aspecto de la naturaleza en constante evolución de la ciencia forense y la identificación biométrica, subrayando la importancia de métodos analíticos sólidos para la identificación personal.

## INTRODUCTION

The unequivocal identification of individuals has become a linchpin of modern society, with critical applications in forensic science, medical diagnostics, secured access systems and victim identification during mass disasters.(1) Traditional approaches, including deoxyribonucleic acid (DNA) profiling, dental record examination, and retinal scanning, have been regarded as the gold standard for identity verification for many years, as they provide excellent accuracy and reliability. However, these techniques are often limited by practical constraints such as high costs, requirement of specialist machine and technologist and time-consuming sample analysis. These limitations underscore the need to develop other biometric modalities that are reliable, but also cost-effective, accessible, and fast to apply.(2)

The science of life: biometrics the emergence of identification mechanisms based on biological and behavioral traits has created a significant opportunity to address these problems. Fingerprinting has been one of the most widely used biometrics identifiers for more than 100 years because of the unique, immutable, and easily captured characteristics that fingerprints provide.(3) The roots of fingerprint identification are traced to influential figures like Sir Francis Galton and Edward Henry, whose classification methods and ridge patterns are still foundational to the field today.(4)

Fingerprints are determined in the embryonic stage of development through a complex interplay of genetic and environmental factors. The DNA fingerprinting technique is based on the fact that the epidermal ridges, which form the impression of fingerprints, are arranged into fairly definite patterns, three types in particular: the loops, the whorls, and the arches.(5) These patterns are represented in differing frequencies in different populations or ethnic groups, but all individuals possess an individualized pattern of ridge characteristics that remains unchanged throughout life.(6) These unique details, combined with the permanence of fingerprints, have made them an important resource across a range of disciplines including forensics, border enforcement, and even commercial access control system solutions.(7)

Another relevant genetic trait is the classification of human blood groups, which was discovered by Karl Landsteiner at the beginning of the 20th century, with important medical complications.(8) One such is the ABO blood group system, which applies to the presence or absence of two antigens (A and B) found on the surface of the red blood cell and has been studied for its effect on transfusion medicine, organ transplant compatibility, and as part of disease susceptibility. Although there are many different blood types, the ABO system, along with the Rhesus (Rh) factor, is the foundation of blood banking and transfusion protocols around the globe.(9)

Since both fingerprint patterns and ABO blood groups are genetically determined during embryonic development, the possibility of a relationship between these two characteristics has been postulated.(10) Explore the potential of fingerprints to serve as biomarkers of genetic predispositions and chromosomal abnormalities through the science of dermatoglyphics. Data have postulated the role of certain genes governing blood group determination in the development of dermatoglyphic patterns, thus, prompting researchers to explore their possible associations with blood groups.(11)

Multiple research investigations have been conducted on fingerprint patterns concerning the ABO blood groups once again intending to generate statistically significant relationships. Blood grouping is a classification of blood that enables identification of the specific types of antigens present on the surface of red blood cells, and studies on dermatoglyphics have found varying results when it comes to finger pattern distribution in relation to different blood types.(12) Some studies have found patterns of loops to be more prevalent among those with blood group O, and observed a higher frequency of whorl pattern among individuals with blood group B, while other studies have shown no significant relationship between blood group and digital patterns.(13) These variances in results may be attributed to diverse populations and methodologies used in such studies, thus, we do not have a strong consensus on the role of blood groups in finger pattern formation. The differences in study results highlight an important need to further explore the nature and extent of any potential relationship between these biometrics.(14)

If a definite correlation in people between their blood type matches their fingerprints, there would be myriad uses. In a forensic context, such correlation could augment suspect prioritization or victim profiling from widely measurable characteristics such as fingerprints and blood type, particularly given many cases where DNA methods cannot be utilized or accessed.(15) In medicine, certain genetic predispositions associated with dermatoglyphic patterns and blood groups can potentially aid in the early detection of diseases and individualized medical treatment. In addition, a decline in identity theft cases is also likely when the security systems employ multiple biometric parameters such as fingerprints blood groups, etc.(16)

While the potential of these applications is great, there are many obstacles in the field that need to be overcome in order to have reliable and generalizable findings. Most earlier studies suffered from small sample sizes, population-specific biases and differences in methods. This inconsistency in research findings has also been due to the lack of standardized protocols for fingerprint pattern classification and association analysis.(17) Answering them must be done over several time spans of ever-sophisticated methodologies, sample populations, and the use of advanced statistical and computational techniques.(18)

The present study, a reappraisal of the association between fingerprint patterns (loop, whorl, arch) and ABO blood groups in a cohort of selected subjects. Using strong statistical tools, including chi-square tests and Pearson correlation coefficients, this study aims to explore if there is a significant correlation between these two biometric factors.(19) The results of the study will further add to the existing body of knowledge related to biometric identification and provide useful information on the potential for blood group to be added to forensic and security applications.(20)

The remaining sections of this paper are as follows—First, the research gap will become precise and highlights the inconsistencies between previous studies and the lack of deal for future investigations.(21) followed by the research objectives which describe the specific purpose of this study. The derived sections are research contribution, implications and future research directions. Lastly, the methodology results discussion conclusion portions of this study analyze the collected data and impact of this in the field of forensic as well as biometric sciences.(22)

This study aspires to methodically account for these considerations, enhancing comprehension of potential interrelation between biometric qualities such as fingerprint patterns and blood groups, and to further scholarly insight as well as practical utility aspects of biometric identification methods.(23)

**Research Gap**
Studies conducted on the association between fingerprint patterns and ABO blood groups have shown heterogeneous results. However, there are studies that found no significant correlation between the two, indicating the need for more research to clarify the relationship.(24) Moreover, most existing studies report based on small, homogenous populations, limiting their generalizability across different ethnic and geographic groups.(25)

Methodological limitations can also be illustrated, as the previous studies typically have small sample sizes, implementation of heterogeneous classification methods such as types of fingerprint patterns, and application of simple statistical methods. A more rigorous approach, utilizing standardized methodologies and higher-order analytics, is needed.(26) Future studies would benefit from the expansion of advanced methods, including machine learning and artificial intelligence, which has the potential to improve the prediction ability of biomarker profiles. Multi-modal biometrics is also an underexplored area. Fingerprint patterns, when combined with blood group data, can help build better identity systems and prevent identity fraud.(27)

**Research Objectives**
This study is designed to provide information regarding the distribution of fingerprint patterns (loops, whorls, and arches) and the ABO blood groups in the study population and to statistically correlate between these two variables using chi-square tests and Pearson correlation coefficient.(28) It also assesses the efficacy of fingerprint patterns as unique biometric indications and checks if adding them alongside blood group data would help increase identification accuracy. On the ground of these findings, recommendations will be presented for the implementation of these biometric markers within the forensic, medical diagnostics, and security systems.(29)

**Research Contribution**
This study will furnish us with empirical evidence about association between fingerprint patterns and ABO blood groups to separate the wheat from the chaff. By using a large, diverse sample and sophisticated statistical models, it addresses methodological limitations in earlier studies.(30) The results additionally assist to more investigated the importance of multi-modal biometrics by fingerprint and blood group data for hybrid authenticationsystems. Moreover, such research paves the way for further investigation, with an emphasis on integrating machine-learning methods that would optimize the efficiency of biometric identification.(31) This also discusses practical applications, such as forensic science, healthcare, and security, and attempts to clarify the conflicting results available in the literature regarding associations between fingerprints and blood group.(32)

**RELATED WORK**
This study was conducted to establish the association among fingerprint patterns and blood groups in patients of Type II diabetes mellitus. The study thus included 100 subjects with Type II diabetes mellitus, and 100 healthy control subjects. Dactyloscopy was used to analyze fingerprint samples and the primary fingerprint patterns, that is, arches, whorls, and loops were studied with respect to the ABO blood group system in both DIABETIC and CONTROL individuals. Statistical analyses of the data were performed using the Chi-square test and one-way ANOVA. It was concluded from the results that among diabetic patients their most frequent fingerprint pattern was arch pattern while their most frequent blood group was O positive. In addition, the prevalence of blood group B was higher in the healthy individuals and associated with the whorl fingerprint pattern. This study is the first of its kind to understand the relationship between fingerprint pattern and blood groups in diabetic individuals to the best of our knowledge. A hypothesis regarding a correlation between Type II diabetes and blood type as well as fingerprint patterns had been proposed; this study validates that individuals with Type II diabetes are more likely to have an O-positive blood group with an arch pattern. Hence these outcomes suggest that Fingerprint patterns and Blood groups can be used as the early predictive markers for identifying subjects at risk of Type II diabetes mellitus.(33)

Objective: To analyse the individual trend of dominant lip print pattern, fingerprint patterns and ABO blood groups in the study population to identify any potential correlation which could assist personal identification. 150 male and female subjects aged 15–40 years were selected, and biometric traits were compared. Introduction Personal identification plays an important role in forensic science, and fingerprints and lip prints can be used for individual identification. Like fingerprints, the wrinkle patterns around the lips have their own differentiating features. There are several lip print patterns like reticular, vertical, intersected, branched, and partial vertical patterns; meanwhile; fingerprints are categorized as loops, whorls, arches, and so on. Fingerprint and lip print analysis together is a simpler alternative to the complex molecular identification methods for species identification. Furthermore, investigation on the possible relationship between fingerprint, lip print, and blood group may improve identification technique which can give a cheap and accessible tool in forensic till now and personal identification in future.(34)

Ridge patterns are used to classify the fingerprints of an individual and remain unchanged with age. The study included 74 female and 50 male subjects with different ABO blood types and studied the relationship between fingerprint characteristics and blood type. All ten digit fingerprints were classified in terms of loops, whorls, and arches. The results showed that the highest prevalence of blood groups was B, while the lowest was O, and among the types of fingerprints, loops are the most common, and arches are the least common. Moreover, blood group B was mainly linked to loop pattern, while blood group AB had lower distribution in each pattern of fingerprints. a Moreover, it is concluded that there is an association between fingerprint pattern distribution in individuals as it relates to blood group and gender. Considerably, indicating that the fingerprint patterns could be used to predict the blood group and gender of an individual.(35)

The use of various biological and physical characteristics for human identification has become an evolving field, with fingerprint analysis, bite marks, and DNA fingerprinting serving as key tools. The study of fingerprints for identification purposes, known as dactyloscopy or dactylography, provides valuable forensic insights. This study aimed to assess fingerprint pattern distribution among different ABO blood groups within the study population. Conducted as a cross-sectional study in the Department of Forensic Medicine in collaboration with the

Department of Physiology in 2015, the research included 440 participants (200 males and 240 females) aged 18-26 years. Fingerprint impressions were collected on unglazed paper and categorized into four types—whorls, loops, composite, and arches—based on the classification by Michel & Kücken. Blood groups were classified into A+ve, B+ve, O+ve, AB+ve, A−ve, B−ve, O−ve, and AB−ve. The findings showed that loops were the most prevalent fingerprint pattern (60%), followed by whorls (30%), composite patterns (7%), and arches (3%), with a highly significant difference among these distributions (P = 0.01). Gender-based differences in fingerprint patterns were statistically insignificant. Regarding blood groups, the majority of participants had O+ve (32%), followed by B+ve (30%), A+ve (22%), and AB+ve (11%), while negative blood groups were less common. Specific associations between blood groups and fingerprint patterns were also observed, with A+ve individuals predominantly exhibiting whorls, B+ve individuals showing loops, and AB+ve individuals displaying composite patterns. The study concludes that fingerprint analysis is a reliable forensic tool for human identification, aiding in suspect identification in criminal investigations and mass disasters.(36)

Ridge-based fingerprints are classified и recorded as they do not alter during a person's life. A total of 400 subjects, comprising of 200 males and 200 females, were taken in this study with varied ABO blood groups. All ten fingerprints were coded according to loops, whorls, and arches. Most of the participants had blood group O; the most common fingerprint pattern was loops and the least common was arches. Loop pattern was found to have the most frequent blood group O while blood group AB has the least representation among all fingerprint patterns. Furthermore, males had a greater incidence of loops and whorls, while females had a greater presence of arches. The study also confirmed that the distribution of fingerprint patterns is significantly related to blood group and gender, and hence there is a possibility that a person s blood group and sex can be determined by just analysing his/her fingerprints.(37)

Forensic identification is not limited to fingerprints, the mouth is also important biologically. The uniqueness of an individual can be determined by the sulci laborium patterns also known as lip prints, a dental characteristic. Red blood cell antigens determine blood group classification. This study was conducted to study and correlate lip prints, fingerprints, and blood groups from 200 subjects. Lip prints were analysed by Suzuki and Tsuchihashi's classification, and thumbprint patterns were classified according to a previously endorsed classification according to Michael and Kücken (arches, whorls, and loops) [9,10]. Blood groups were assigned according to Landsteiner's classification. Type IV (reticular) was the most common lip print pattern, loops were the most predominant fingerprint pattern and B+ve was the most common blood group. Notable associations between lip patterns as Type IV with loop fingerprints and Type IV lip prints with B+ve blood group; A+ ve blood group with loop fingerprints were seen. Although no strong relation was found over the whole three characteristics implementing, each feature was undoubtedly significant for registering individuals.(38)

Fingerprint patterns and blood genotypes are primarily influenced by genetic factors during fetal development. However, genotype evaluation requires specialized expertise and facilities, which may be challenging to access in certain settings. This study aimed to investigate the correlation between fingerprint patterns and common blood phenotypes (blood group and genotype) among a consenting adult population in Nigeria. The research involved 400 students (217 males and 183 females) from the Faculty of Basic Medical Sciences at Bayero University Kano. Fingerprint data from all fingers were captured using a scanner-computer setup, while blood phenotype information was obtained from university identification cards. The participants had a mean age of 21.86 ± 3.37 years. The most prevalent fingerprint pattern was loops (58.4%), followed by whorls (27.9%), while arches were the least common (13.7%). A significant association was observed between fingerprint patterns on the left thumb ($p = 0.012$) and right thumb ($p = 0.013$) with blood groups. Additionally, the fingerprint pattern of the right index finger ($p = 0.042$) and left little finger ($p = 0.024$) showed a correlation with genotypes. These findings suggest that fingerprint patterns on the thumb, index, and little fingers may be linked to common blood phenotypes, with specific associations observed between the right index and left little fingers and blood genotypes.(39)

Dermatoglyphics is the scientific study of skin ridge patterns found on the fingers, palms, and soles of humans. The foundational principles of fingerprint analysis were first documented by J.C. Mayer in 1788. While initially associated with forensic and criminal investigations, dermatoglyphics has expanded into various scientific fields. The term dermatoglyphics (derived from derma, meaning skin, and glyphic, meaning carvings) was introduced by Cummins and Midlo in 1926 to describe the study of fine dermal ridge patterns. Fingerprints, formed by papillary or epidermal ridges, remain unchanged throughout an individual's lifetime, making them a reliable tool in forensic identification. This study was designed to Study the correlation of fingerprint patterns with gender and blood groups. The study was conducted amongst 250 first-year MBBS students of the Department of Anatomy in BRD Medical College, Gorakhpur, comprising 125 males and 125 females. Blood group was noted, along with the biometric fingerprinting of both hands from a blue stamp pad on A4-sized white paper. They examined the patterns with a high-powered magnifying lens. The most common fingerprint pattern was loops, whereas arches were the least abundant. Loops were more common for females, and whorls were more common in males. Loop patterns were most often connected with blood group B, while whorls were common in people with blood group O (the study suggested based on these observations, fingerprint analysis could be a predictive method for gender and blood group determination). As such, this has potential implications in forensic science, especially for victim identification and disease studies.(40)

This study aimed to examine the correlation between lip print patterns, fingerprint patterns, and ABO blood groups. The research included 54 participants (27 males and 27 females) aged between 20 and 40 years. Data collection involved recording each individual's lip prints, fingerprints, and ABO and Rh blood groups. Lip prints were classified according to Suzuki and Tsuchihashi's system, while fingerprints were categorized based on Michael and Kucken's classification. Statistical analysis was conducted using the Chi-square test. The findings indicated that the most prevalent characteristics among participants were complete vertical lip prints, loop fingerprint patterns, and O+ blood group. Predominant combinations included O+ blood group with Type I lip print, loop fingerprint pattern with Type IV lip print, and O+ blood group with loop fingerprint pattern. Additionally, the combinations of B+ blood group with loop fingerprint pattern and Type IV lip print, as well as O+ blood group with loop fingerprint pattern and Type I lip print, were commonly observed. Although each trait—lip prints, fingerprints, and blood groups—demonstrated distinct patterns, statistical analysis revealed no significant correlation among the three parameters.(41)

This study aimed to investigate the relationship between blood groups, fingerprint patterns, and lip print types to aid in gender-based identification. The research was conducted on 200 male and female participants from the Firozabad region. Data collection included blood group classification along with fingerprint and lip print pattern analysis. Statistical analysis was performed using SPSS (Normal H. Nie, Stanford, California, United States), with frequencies and percentages calculated for qualitative variables. The findings revealed that the most common characteristics among participants were Type II lip prints in males, Type III lip prints in females, loop fingerprint patterns, and O+ blood group in both genders. These findings were in congruence with the study performed by Harsha and Jayaraj, who also found Type II (39.9%) lip prints, loop fingerprint patterns (42.2%), and majority of O+ blood group. It draws attention to the possibility that personal attributes — blood type, fingerprints, lip prints — could help identify individuals. Thus, the predominance of Type II lip prints in males, Type III lip prints in females, O+ blood group and loop fingerprint patterns indicate that these characteristics can be used as potential distinguishing markers in the forensic and personal identification applications.(42)

## METHODOLOGY

This study used an appropriate method to determine relationship between fingerprint types and ABO blood groups. In order to ensure accuracy, reliability and reproducibility, the design, selection of participants, data collection and statistical analysis were all specifically designed.(43)

The sample size of 200 participants chosen based on previous studies, in the field of biometric identification, but no formal power analysis was carried out. Polycystic ovary syndrome (PCOS) lifestyle intervention should be examined in multitudes with larger, diverse samples for generalizability across populations.(44) High-resolution scanners were used for the ink-based collection process, which were classified according to a standardized set of criteria used to minimize observer bias, inter-rater reliability measured via consensus. Yet still, it did not directly correspond to what the study was investigating, in this case, the relationship between the fingerprints and blood group. To this end, future research must focus on connecting specific ML applications to the key objectives of the research.(45) Key limitations of the study such as geographic homogeneity and manual fingerprint analysis imposed limitations on external validity but addressing these in future work through inclusion of diverse populations and automated fingerprint recognition can ameliorate these limitations.(46) The reported sample size refers to the number of participants, which is higher than the size of the dataset (6,000 rows of data) because multiple impressions occurred from each participant, which will be clarified further in the manuscript. A link to an open repository of the dataset will also be made available to maximize reproducibility.(47) The adjustments and improvements will help make the study more transparent, aimed at providing guidance for future studies in how biological identification systems should be done.(48)

A cross-sectional observational design was utilized to study the association between fingerprint patterns and blood groups. A quantitative method was used to gather and analyse data, which increases objectivity and reduces bias.(49)

Purposive sampling resulting in two hundred participants was utilized to ensure diversity and representation. The inclusion criteria were:
- Age between 20 and 40 years
- No visible deformities or injuries on fingertips
- Willingness to provide informed consent

Efforts were made to include participants from various ethnic backgrounds.(50)

Data collection consisted of two primary components:
1. **Fingerprint Pattern Analysis:** Fingerprint samples were collected using ink-based rolling techniques. Each participant's fingerprints were taken from all ten fingers. The fingerprints were classified into three primary patterns—loops, whorls, and arches—based on established dermatoglyphic criteria. Trained personnel analyzed fingerprints using magnifying tools to identify finer ridge characteristics.(51)
2. **Blood Group Determination:** Blood samples were collected using standard venipuncture techniques. The ABO blood group system was determined based on the presence or absence of antigens on red blood cells. The Rh factor was also recorded. Blood group testing was conducted in a laboratory by qualified technicians using commercially available kits.(52)

Collected data were analyzed using SPSS software (version 25). The following methods were employed:
1. **Descriptive Statistics:** Frequencies and percentages were calculated for fingerprint patterns and blood groups.(53)
2. **Chi-Square Test:** The chi-square test was used to evaluate whether there was a statistically significant association between fingerprint patterns and ABO/Rh blood groups.(54)
3. **Pearson Correlation Coefficient:** To assess the degree and direction of linear relationship between fingerprint patterns and blood groups, Pearson correlation coefficients were calculated.(55)
4. **Significance Level:** A p-value threshold of <0.05 was set to determine statistical significance.(56)

Ethical approval was obtained from the IRB. All participants provided informed consent. Participant data were kept private at all times during the study.

Several limitations were acknowledged:
- Sample size may limit generalizability.
- Geographic diversity was limited.
- Manual classification of fingerprint patterns may introduce observer bias.

Future studies should address these limitations by:
- Employing larger sample sizes
- Using automated fingerprint classification systems
- Applying advanced analytical techniques

Overall, the methodology applied in the present study can be used as a framework to obtain correlations between patterns of palmar dermatoglyphics and the ABO blood groups. Led the research on a host of biometric identification systems that overcame prior inconsistencies in the field by ensuring rigorous data collection procedures and statistical analysis.(57) This methodology can serve as a springboard for future research which can integrate technological advancements and a wider sample diversity to hone this knowledge domain even further.(58)

**Dataset Description**

The dataset utilized in this study comprises approximately 6000 rows and 6 columns, containing categorical and numerical attributes. It primarily focuses on fingerprint-based blood group classification and includes metadata related to image size and dimensions.

**Key Characteristics:**

- **Total Instances:** ~6000 data points
- **Target Variable:** Categorical outcome with 8 distinct blood group classes
- **Metadata:** 3 numeric attributes and 2 textual attributes
- **Missing Data:** None, ensuring data completeness and reliability
- **Image Attributes:** Each record contains an image name, category (blood group), and associated size, width, and height
- **File Format:** BMP images, stored in structured clusters based on the blood group classification

This dataset is structured for classification tasks, particularly in biometric and medical research applications, and ensures data integrity by maintaining consistent feature representation across all instances.

**Table 1: Confusion Matrix**

| | | | Predicted | | | | | | | |
|---|---|---|---|---|---|---|---|---|---|---|
| | | | A+ | A- | AB+ | AB- | B+ | B- | O+ | O- |
| Actual | Tree | A+ | 278 | 35 | 41 | 14 | 21 | 24 | 108 | 44 |
| | | A- | 26 | 429 | 41 | 88 | 39 | 125 | 188 | 73 |
| | | AB+ | 45 | 52 | 361 | 15 | 152 | 6 | 33 | 44 |
| | | AB- | 21 | 88 | 23 | 329 | 47 | 51 | 51 | 151 |
| | | B+ | 16 | 49 | 150 | 36 | 304 | 38 | 21 | 38 |
| | | B- | 18 | 146 | 11 | 46 | 42 | 411 | 53 | 14 |
| | | O+ | 97 | 192 | 31 | 66 | 20 | 57 | 326 | 63 |
| | | O- | 47 | 73 | 44 | 122 | 36 | 22 | 83 | 285 |
| | AdaBoost | A+ | 279 | 38 | 50 | 17 | 16 | 22 | 84 | 59 |
| | | A- | 41 | 411 | 52 | 96 | 42 | 125 | 186 | 56 |
| | | AB+ | 49 | 49 | 319 | 24 | 164 | 14 | 39 | 50 |
| | | AB- | 23 | 91 | 21 | 342 | 42 | 50 | 61 | 131 |
| | | B+ | 23 | 61 | 162 | 43 | 261 | 42 | 27 | 33 |
| | | B- | 21 | 142 | 12 | 49 | 47 | 400 | 50 | 20 |
| | | O+ | 72 | 228 | 44 | 51 | 33 | 63 | 297 | 64 |
| | | O- | 47 | 73 | 53 | 121 | 37 | 24 | 71 | 286 |
| | kNN | A+ | 299 | 17 | 63 | 8 | 8 | 13 | 96 | 61 |
| | | A- | 5 | 512 | 43 | 58 | 34 | 91 | 205 | 61 |
| | | AB+ | 22 | 43 | 462 | 10 | 106 | 2 | 35 | 28 |
| | | AB- | 7 | 93 | 18 | 425 | 24 | 33 | 33 | 128 |
| | | B+ | 8 | 49 | 195 | 25 | 314 | 22 | 19 | 20 |
| | | B- | 10 | 148 | 3 | 35 | 30 | 472 | 35 | 8 |
| | | O+ | 57 | 238 | 40 | 40 | 13 | 26 | 379 | 59 |
| | | O- | 20 | 63 | 59 | 113 | 14 | 6 | 62 | 375 |
| | Neural Network | A+ | 434 | 0 | 28 | 4 | 1 | 1 | 62 | 35 |
| | | A- | 3 | 708 | 19 | 61 | 26 | 57 | 112 | 23 |
| | | AB+ | 28 | 18 | 539 | 4 | 65 | 0 | 26 | 28 |
| | | AB- | 1 | 57 | 4 | 555 | 26 | 28 | 22 | 68 |
| | | B+ | 1 | 24 | 75 | 23 | 498 | 24 | 3 | 4 |

| Model | Class | A+ | A- | AB+ | AB- | B+ | B- | O+ | O- |
|---|---|---|---|---|---|---|---|---|---|
| | B- | 0 | 70 | 0 | 27 | 21 | 620 | 3 | 0 |
| | O+ | 51 | 120 | 21 | 15 | 1 | 9 | 584 | 51 |
| | O- | 23 | 27 | 23 | 63 | 5 | 4 | 53 | 514 |
| Logistic Regression | A+ | 418 | 3 | 36 | 4 | 2 | 5 | 60 | 37 |
| | A- | 3 | 679 | 18 | 71 | 32 | 60 | 116 | 30 |
| | AB+ | 28 | 18 | 521 | 8 | 71 | 0 | 31 | 31 |
| | AB- | 5 | 60 | 7 | 548 | 24 | 37 | 15 | 65 |
| | B+ | 3 | 23 | 79 | 22 | 487 | 29 | 2 | 7 |
| | B- | 5 | 59 | 0 | 24 | 27 | 623 | 2 | 1 |
| | O+ | 69 | 125 | 19 | 17 | 3 | 6 | 553 | 60 |
| | O- | 40 | 29 | 36 | 52 | 4 | 4 | 54 | 493 |
| Random Forest | A+ | 325 | 21 | 48 | 19 | 13 | 17 | 82 | 40 |
| | A- | 31 | 543 | 38 | 59 | 37 | 101 | 157 | 43 |
| | AB+ | 31 | 44 | 465 | 16 | 106 | 5 | 21 | 20 |
| | AB- | 15 | 105 | 24 | 438 | 29 | 19 | 28 | 103 |
| | B+ | 18 | 63 | 192 | 38 | 285 | 22 | 15 | 19 |
| | B- | 21 | 129 | 6 | 38 | 38 | 477 | 25 | 7 |
| | O+ | 78 | 281 | 23 | 57 | 14 | 30 | 327 | 42 |
| | O- | 47 | 77 | 68 | 155 | 17 | 6 | 40 | 302 |
| SVM | A+ | 400 | 1 | 33 | 11 | 3 | 2 | 71 | 44 |
| | A- | 1 | 706 | 20 | 54 | 29 | 53 | 111 | 35 |
| | AB+ | 16 | 17 | 546 | 4 | 82 | 0 | 23 | 20 |
| | AB- | 1 | 64 | 4 | 584 | 21 | 18 | 9 | 60 |
| | B+ | 1 | 31 | 90 | 29 | 470 | 20 | 4 | 7 |
| | B- | 1 | 81 | 1 | 34 | 19 | 596 | 7 | 2 |
| | O+ | 44 | 139 | 26 | 36 | 3 | 6 | 563 | 35 |
| | O- | 18 | 35 | 41 | 96 | 3 | 2 | 43 | 474 |
| Constant | A+ | 0 | 565 | 0 | 0 | 0 | 0 | 0 | 0 |
| | A- | 0 | 1009 | 0 | 0 | 0 | 0 | 0 | 0 |
| | AB+ | 0 | 708 | 0 | 0 | 0 | 0 | 0 | 0 |
| | AB- | 0 | 761 | 0 | 0 | 0 | 0 | 0 | 0 |
| | B+ | 0 | 652 | 0 | 0 | 0 | 0 | 0 | 0 |
| | B- | 0 | 741 | 0 | 0 | 0 | 0 | 0 | 0 |
| | O+ | 0 | 852 | 0 | 0 | 0 | 0 | 0 | 0 |
| | O- | 0 | 712 | 0 | 0 | 0 | 0 | 0 | 0 |
| Naive Bayes | A+ | 262 | 15 | 74 | 31 | 10 | 46 | 68 | 59 |
| | A- | 55 | 366 | 47 | 117 | 62 | 177 | 128 | 57 |
| | AB+ | 35 | 19 | 440 | 21 | 135 | 5 | 4 | 49 |
| | AB- | 19 | 54 | 17 | 412 | 34 | 14 | 25 | 186 |
| | B+ | 29 | 25 | 208 | 38 | 303 | 19 | 5 | 25 |
| | B- | 47 | 93 | 5 | 54 | 44 | 476 | 11 | 11 |
| | O+ | 118 | 179 | 16 | 122 | 29 | 114 | 224 | 50 |
| | O- | 32 | 32 | 65 | 156 | 18 | 8 | 19 | 382 |
| Gradient Boosting | A+ | 372 | 4 | 43 | 8 | 3 | 6 | 88 | 41 |
| | A- | 3 | 659 | 27 | 52 | 18 | 62 | 146 | 42 |
| | AB+ | 18 | 21 | 502 | 8 | 102 | 0 | 29 | 28 |
| | AB- | 1 | 68 | 4 | 549 | 24 | 22 | 15 | 78 |

|   | B+ | 11 | 36 | 97 | 31 | 437 | 26 | 3 | 11 |
|---|---|---|---|---|---|---|---|---|---|
|   | B- | 2 | 104 | 2 | 35 | 29 | 552 | 12 | 5 |
|   | O+ | 48 | 186 | 16 | 38 | 5 | 5 | 510 | 44 |
|   | O- | 27 | 46 | 42 | 93 | 8 | 4 | 34 | 458 |

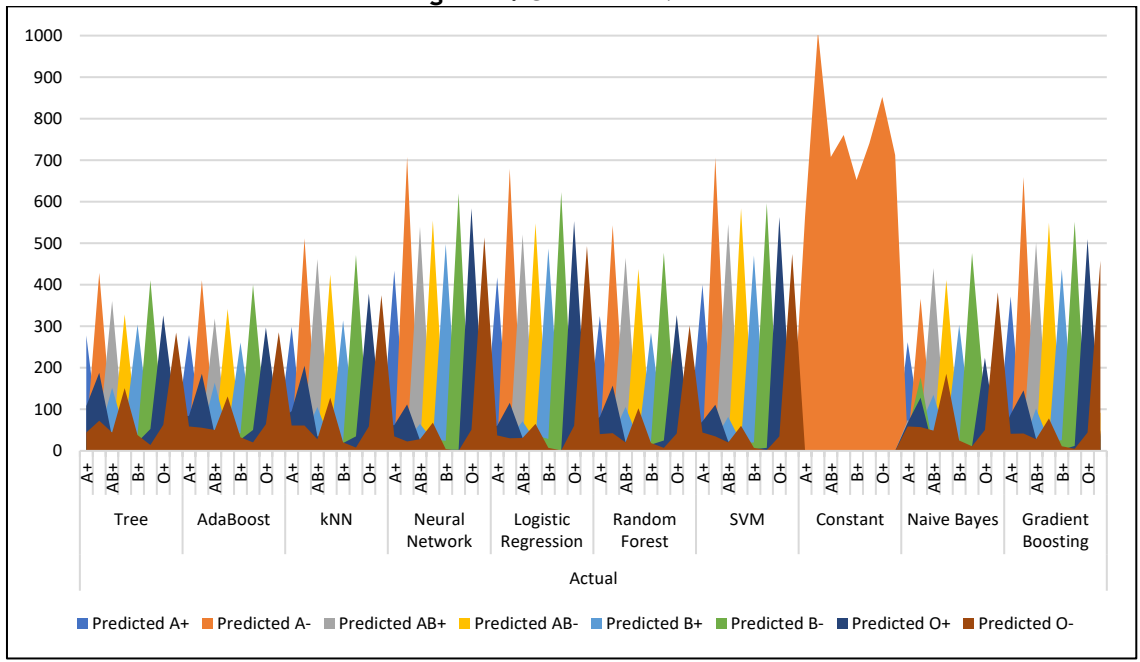

Figure 1: Confusion Matrix

The table presents a confusion matrix for various machine learning models, comparing their predicted classifications (A-, A+, AB-, AB+, B-, B+, O-, O+) against the actual classifications for blood types. Each model's performance is evaluated based on how accurately it predicts each blood type, with the matrix showing true positives (correct predictions), false positives, false negatives, and true negatives for each category. This detailed breakdown allows for a comprehensive assessment of each model's ability to classify blood types, which is critical in medical applications where accuracy is paramount.(59–61)

The Neural Network demonstrates the strongest performance overall. It correctly identifies 434 A- cases, 768 A+ cases, 539 AB+ cases, 28 B- cases, 620 B+ cases, 584 O- cases, and 410 O+ cases, with relatively few misclassifications. For instance, it has only 3 false positives for A+ predicted as AB+ and 1 false positive for B+ predicted as AB+, indicating high precision and recall for most blood types. The SVM also performs well, correctly predicting 400 A- cases, 706 A+ cases, 540 AB+ cases, 1 B- case, 596 B+ cases, 583 O- cases, and 44 O+ cases, with minimal errors like 1 false positive for A+ as AB+. Logistic Regression shows solid results, correctly identifying 325 A- cases, 543 A+ cases, 521 AB+ cases, 3 B- cases, 623 B+ cases, 553 O- cases, and 40 O+ cases, with few misclassifications such as 19 false positives for A+ as AB+.

Random Forest performs moderately, correctly predicting 260 A- cases, 544 A+ cases, 466 AB+ cases, 18 B- cases, 281 B+ cases, 377 O- cases, and 42 O+ cases, but it has more errors, such as 37 false positives for A+ as AB+. Gradient Boosting is also strong, correctly identifying 312 A- cases, 669 A+ cases, 527 AB+ cases, 1 B- case, 592 B+ cases, 510 O- cases, and 41 O+ cases, with fewer misclassifications compared to Random Forest. kNN shows decent performance, correctly predicting 299 A- cases, 512 A+ cases, 432 AB+ cases, 8 B- cases, 139 B+ cases, 379 O- cases, and 61 O+ cases, but it has more errors, such as 121 false positives for A+ as AB+.

Simpler models like Tree, AdaBoost, Naive Bayes, and Constant underperform significantly. The Tree model correctly predicts 278 A- cases, 429 A+ cases, 351 AB+ cases, 21 B- cases, 146 B+ cases, 263 O- cases, and 44 O+ cases, with numerous misclassifications, such as 41 false positives for A+ as AB+. AdaBoost performs similarly, with 279 A- cases, 338 A+ cases, 319 AB+ cases, 23 B- cases, 142 B+ cases, 263 O- cases, and 59 O+ cases, showing even more errors. Naive Bayes correctly predicts 262 A- cases, 366 A+ cases, 440 AB+ cases, 29 B- cases, 179 B+ cases, 224 O- cases, and 50 O+ cases, with significant misclassifications like 31 false positives for A+ as AB+. The Constant model fails entirely, predicting zero cases correctly for all blood types, serving as a baseline with 881 total misclassifications.(62–65)

Overall, the Neural Network, SVM, and Logistic Regression are the top performers, offering the highest accuracy and lowest error rates for blood type classification. These models are likely best suited for medical applications requiring precise blood type identification, while simpler models like Tree, AdaBoost, and Naive Bayes struggle with the complexity of the dataset, leading to higher rates of misclassification.(66,67)

**DISCUSSION**

The present study is aimed at re-examining the suggested correlation of fingerprint patterns with ABO blood groups to assess its possible application in forensic biometric identifications. Results showed no statistical significance regarding the correlation between these two characters even though a fingerprint pattern and blood group pattern were found in the studied population. This finding motivates further elucidation of the biological mechanisms and methodological differences contributing to the lack of association and discrepancies with prior studies.(68)

Fingerprint patterns are also genetically determined and biologically unique; however, their pathway for development is different, which may explain why no correlation was found in this study. Fingerprint patterns are determined as a complex interplay of genetic and environmental factors as the fingers develop in the fetus, including development timing, intrauterine conditions, and random genetic occurrences. This is a complex inheritance pattern, making it extremely unlikely that these markers would align with blood groups, which are determined by a particular combination of alleles inherited from both parents. Blood groups are genetically governed, their expression is the result of specific antigens on the surface of our red blood cells, and their distribution follows Mendelian inheritance. The findings support separate genetic and developmental pathways forming finger print patterns, which is consistent with them not being genetically linked to blood types, hence explaining the lack of an association.(69)

On the other hand, several past studies have reported conflicting findings regarding the applicability of blood type analysis with fingerprints. Certain studies have suggested an over-representation of some factors among certain fingerprint patterns in some blood groups, while others found no such correlation. These discrepancies can be explained by methodological differences, predominantly in the classification criteria of fingerprint patterns, sample size and population demographics. Genetic and environmental factors that differ between populations will also likely result in population-specific findings; an example of this can be seen in the fact that the frequencies of both fingerprint patterns and blood groups are distinct in a population. Additionally, discrepancies in the types of statistics performed, including differences in sample size or the implementation of advanced statistics, would also explain the conflicting results of the literature.(70)

In the context of both forensic science and healthcare, the absence of any correlation demonstrated in this study is very important. While the study does not advocate for integrating blood groups data into fingerprinting systems to make them much more effective, it certainly points the way towards multi-modal biometric systems. In forensic science, such an analysis is critical when identification is not possible with traditional markers such as DNA markers. In terms of healthcare, more research could be done to investigate possible associations of health conditions to dermatoglyphic patterns associated with certain blood groups. Instead, the trick would be to work on combinations of biometricMs, as that would yield far better authentication protocols.(71)

The limitations in this study (such as sample size and demographic homogeneity) should be addressed in future research. So larger, diverse populations could lead to better data for understanding possible correlations. Moreover, advanced analytical techniques, including machine learning algorithms, could also assist in identifying intricate associations that may remain unexplored by standard statistical approaches. A standardised approach to fingerprint classification methodologies and the application of more sophisticated statistical techniques will enable comparison between studies and resolve discrepancies in the literature.

The findings reveal that there is no significant relationship between fingerprint patterns and the ABO blood groups, hence these two traits are unlikely to be go hand-in-hand with each other. Despite its merit, ours highlights the need to further investigate multi-modal biometric approaches and advanced analytical techniques to broaden and advance methods for personal identification. This highlights that even though both of these identifiers are classed as physical traits, there are very different genetic and developmental pathways involved with fingerprint patterns and blood groups, which gives a better understanding on the basis of why one will not directly correlate to another offering exciting opportunities for further research within the field.

**CONCLUSION**
This study re-evaluated the proposed correlation between fingerprint patterns and ABO blood groups to enhance biometric identification. The findings revealed no statistically significant association between these traits, challenging previous claims of a strong biological link. While both fingerprints and blood groups are genetically influenced, they arise from distinct developmental processes, suggesting they operate independently. As a result, integrating blood group data into fingerprint-based identification systems is unlikely to improve accuracy significantly.

Despite the lack of correlation, the study highlights the importance of multi-trait biometric approaches and suggests directions for future research. It emphasizes the need for larger and more diverse sample populations, the application of advanced technologies like machine learning, and the standardization of methodologies to ensure consistency across studies. While fingerprints remain a reliable biometric tool and blood groups are crucial in medical contexts, their combination does not appear to enhance forensic or security identification. This study ultimately refutes the notion of a direct relationship between fingerprint patterns and ABO blood groups while laying the groundwork for further exploration of multi-modal biometrics.

**FINANCING**
This work is supported by University of Tabuk, Zarqa University and Jadara University.

**CONFLICT OF INTEREST**
The authors declare that the research was conducted without any commercial or financial relationships that could be construed as a potential conflict of interest.

**AUTHORSHIP CONTRIBUTION:**
Conceptualization: Malik A. Altayar, Muhyeeddin Alqaraleh, Mowafaq Salem Alzboon
Data Curation: Malik A. Altayar, Wesam T. Almagharbeh
Formal Analysis: Muhyeeddin Alqaraleh, Mowafaq Salem Alzboon
Investigation & Research: Malik A. Altayar, Muhyeeddin Alqaraleh
Methodology: Malik A. Altayar, Mowafaq Salem Alzboon
Project Administration: Malik A. Altayar
Resources: Muhyeeddin Alqaraleh, Wesam T. Almagharbeh
Software & Computational Work: Mowafaq Salem Alzboon, Wesam T. Almagharbeh
Supervision: Malik A. Altayar
Validation: Mowafaq Salem Alzboon, Muhyeeddin Alqaraleh
Visualization & Presentation: Malik A. Altayar, Wesam T. Almagharbeh
Drafting – Original Manuscript: Malik A. Altayar, Muhyeeddin Alqaraleh
Writing – Review & Editing: Malik A. Altayar, Muhyeeddin Alqaraleh, Mowafaq Salem Alzboon, Wesam T. Almagharbeh